\documentclass[10pt,journal,compsoc]{IEEEtran}
\ifCLASSOPTIONcompsoc
  \usepackage[nocompress]{cite}
\else
  \usepackage{cite}
\fi
\ifCLASSINFOpdf
\else
\fi
\usepackage{amsmath}
\usepackage[cmintegrals]{newtxmath}

\long\def\comment#1{}
\usepackage{xcolor}

\newcommand{\ie}{{\it i.e.}}
\newcommand{\eg}{{\it e.g.}}

\newcommand{\x}{\mathbf{x}}
\newcommand{\y}{\mathbf{y}}

\newcommand{\W}{\mathbf{W}}

\usepackage{multirow}
\usepackage{booktabs} 
\usepackage{graphicx}
\usepackage{graphics}
\def\yanbo#1{\textcolor{black}{#1}}

\begin{document}
%
\title{Tencent ML-Images: A Large-Scale Multi-Label Image Database for Visual Representation Learning}

%
%
%

\comment{
\author{Baoyuan Wu,~\IEEEmembership{Member,~IEEE,}
        Weidong Chen,
        Wei Liu,~\IEEEmembership{Sensior Member,~IEEE,}
        Yanbo Fan, ~\IEEEmembership{Member,~IEEE,}
        Yong Zhang, ~\IEEEmembership{Member,~IEEE,}
        Jinlong Hou, ~\IEEEmembership{Member,~IEEE,}
        Junzhou Huang, ~\IEEEmembership{Member,~IEEE,}
        Tong Zhang, ~\IEEEmembership{Fellow,~ASA and IMS}
\thanks{All authors are with Tencent AI Lab, Shenzhen, China. 
B. Wu and W. Chen contribute equally, and B. Wu is the corresponding author, email: wubaoyuan1987@gmail.com. }
}
}

\author{Baoyuan Wu$^\dagger$, 
        Weidong Chen$^\dagger$,
        Yanbo Fan$^\dagger$,  
        Yong Zhang$^\dagger$,  
        Jinlong Hou$^\dagger$, \\ 
        Jie Liu$^\dagger$, 
        Tong Zhang$^\ddagger$ 
\thanks{$^\dagger$ Tencent AI Lab, Shenzhen, China.}
\thanks{$^\ddagger$ Department of Computer Science and Mathematics, Hong Kong University of Science and Technology.}
\thanks{Baoyuan Wu and Weidong Chen contribute equally.}
\thanks{Yanbo Fan is the corresponding author, email: fanyanbo0124@gmail.com.}
\thanks{This work is accepted to IEEE Access, DOI: 10.1109/ACCESS.2019.2956775.}
}

\IEEEtitleabstractindextext{%

\begin{abstract}
	In existing visual representation learning tasks, deep convolutional neural networks (CNNs) are often trained on images annotated with single tag, such as ImageNet. 
	However, single tag annotation cannot describe all important contents of one image, and some useful visual information may be wasted during training. 
	In this work, we propose to train CNNs from images annotated with multiple tags, to enhance the quality of visual representation of the trained CNN model. 
	To this end, we build a large-scale multi-label image database with 18M images and 11K categories, dubbed {\it Tencent ML-Images}. 
	We efficiently train the ResNet-101 model with multi-label outputs on Tencent ML-Images, taking 90 hours for 60 epochs, based on a large-scale distributed deep learning framework, \ie, TFplus. 
	The good quality of the visual representation of the Tencent ML-Images checkpoint is verified through three transfer learning tasks, including single-label image classification on ImageNet and Caltech-256,  object detection on PASCAL VOC 2007, and semantic segmentation on PASCAL VOC 2012.
	The Tencent ML-Images database, the checkpoints of ResNet-101, and all the training codes have been released at {\it https://github.com/Tencent/tencent-ml-images}. It is expected to promote other vision tasks in the research and industry community.
\end{abstract}

}

\IEEEpeerreviewmaketitle

\maketitle

\section{Introduction} 
\label{sec: introduction}

This work presents the large-scale visual representation learning on a newly built multi-label image database, dubbed {\it Tencent ML-Images}. 
We start from the discussions of the following two questions.  
\begin{itemize}
	\item Why we need large-scale image database? Deep learning had been in a long trough, until 2012 when AlexNet \cite{alexnet-cnn-nips-2012} shows surprising results on the single-label image classification task of ILSVRC2012 challenge \cite{imagenet-cvpr-2009}. The excellent potential of deep neural networks is released through the large-scale image database ImageNet \cite{imagenet-cvpr-2009}. Besides, the cost of acquiring training data for many visual tasks, such as object detection and semantic segmentation is very high. Due to the insufficient training data, they usually need certain pre-trained model with good visual presentation on other large-scale database (\eg, single-label image classification model trained on ImageNet) as initialization. 
	\item Why we need multi-label image database? As there are multiple objects in most natural images, single-label annotation may miss some useful information and mislead the training of CNNs. For example, two visually similar images that include both {\it cow} and {\it grass} may be annotated as {\it  cow} and {\it grass} separately. The reasonable approach is ``telling'' the CNN model that these two images contain both {\it cow} and {\it grass}. 
\end{itemize}

The above discussions explain why we need large-scale multi-label image database for visual representation learning with deep neural networks.  
However, annotating one image with multiple tags is much more time-consuming than annotating one image with single tag, and it is difficult to control the annotation quality. 
To the best of our knowledge, the largest public multi-label image database is Open Images \cite{openimages,kuznetsova2018open}, which includes about 9 million images.
Recently, Sun et al. \cite{google-JFT300m-iccv-2017} fine-tuned a ResNet-101 model \cite{residual-he-cvpr-2016} that pre-trained on JFT-300M (a multi-label image database with 300 million images), leading to $79.2\%$ top-1 accuracy on the validation set of ImageNet. In contrast, the ResNet-101 model trained on ImageNet from scratch only gives $77.5\%$ top-1 accuracy. 
However, the training on JFT-300M takes 2 months for 4 epochs,
as the training size of JFT-300M is 200 times more than that of ImageNet.
Besides, JFT-300M and its checkpoint are not publicly available. 

In this work, we build a new large-scale multi-label image database, dubbed {\it Tencent ML-Images}.
Instead of collecting new images from Google search or Flickr as did in other databases, 
we collect images from existing image databases, \ie, Open Images \cite{openimages} and ImageNet \cite{imagenet-cvpr-2009}. 
Specifically, we merge their class vocabularies into one unified vocabulary. We further remove rare and redundant categories as well as the corresponding images, and obtain about 18 million images with 11,166 categories. 
We then build semantic hierarchy of the unified vocabulary, according to semantic information extracted from WordNet \cite{wordnet-1998}. We also derive the class co-occurrence among categories, which are then used together with  the constructed semantic hierarchy to augment the annotations, based on the original annotations from Open Images and ImageNet. 

To verify the quality of the built Tencent ML-Images, we conduct large-scale visual representation learning with the popular deep neural network ResNet-101. 
There are two main difficulties in the large-scale representation learning using the multi-label image database, including the severe class imbalance and the long training process. 
To alleviate the side-effect of class imbalance, we design a novel loss function that simultaneously considers weighted cross entropy, adaptive loss weight along the training process and down-sampling of negative training images in each mini-batch.  
To accelerate the training process, we utilize the large-scale distributed deep learning framework, \ie, TFplus, with Message Passing Interface (MPI) and NVIDIA Collective Communications Library (NCCL) \cite{NCCL-2017}. Consequently, the whole training process takes 90 hours of 60 epochs, using 128 GPUs. 
Furthermore, to verify the quality of visual representation of the ResNet-101 model pre-trained on Tencent ML-Images, we conduct transfer learning on three other vision tasks, including single-label image classification, object detection and semantic segmentation. 
We compare with the transfer learning using the checkpoints pre-trained on JFT-300M and ImageNet, respectively. The better transfer learning results using the checkpoint pre-trained on Tencent ML-Images demonstrate the good quality of Tencent ML-Images and the trained checkpoints. 

The main contributions of this work are four-fold.
\begin{itemize}
	\item We build a multi-label image database with 18M images and 11K categories, dubbed {\it Tencent ML-Images}, which is the largest publicly available multi-label image database until now. 
	\item We efficiently train the ResNet-101 model on Tencent ML-Images, utilizing a large-scale distributed deep learning framework. Besides, we design a novel loss function to alleviate the side-effect of the severe class-imbalance in large-scale multi-label database. 
	\item We demonstrate that the good quality of Tencent ML-Images and its pre-trained checkpoint through the transfer learning on three different vision tasks. 
	\item We release the Tencent ML-Images database, the trained ResNet-101 checkpoints, as well as the complete codes of data preperation, pre-training and fine-tuning, at the GitHub address {\it https://github.com/Tencent/tencent-ml-images}. It is expected to promote other vision tasks for the research and industry community. 
\end{itemize}

The organization of this manuscript is as follows. 
Related work is reviewed in section \ref{sec: related work}.
The built multi-label image database is introduced in section \ref{sec: dataset}, including the image source, class vocabulary, semantic hierarchy, tag augmentation and statistic informations. 
The visual representation learning on Tencent ML-Images is presented in section \ref{sec: experiment multi-label}.
Transfer learning to single-label image classification, object detection, and semanic segmentation are presented in section \ref{sec: transfer learning}, followed by the conclusion in section \ref{sec: conclusion}.

\section{Related Work}
\label{sec: related work}

In this section, we review the image databases that are used for visual representation learning. They can be generally partitioned into two categories. One category is the single-label image database, where each image is annotated with only one tag. The other category is the multi-label image database, where each image is annotated with multiple tags. 

Widely used single-label image databases include CIFAR-10 \cite{cifar10-2009}, Caltech-256 \cite{caltech-256}, MNIST \cite{mnist-2010}, ImageNet \cite{imagenet-cvpr-2009}, WebVision \cite{webvision-2017}, SUN \cite{sun-cvpr-2010} and Places \cite{places-2017}, etc. 
Before the deep learning era, the scales of most image databases are not very large.
CIFAR-10 \cite{cifar10-2009} includes 60K small-sized natural images with 10 categories.
Caltech-256 \cite{caltech-256} includes 30,607 images with 256 object categories. 
MNIST \cite{mnist-2010} includes 70K images of handwritten digits, from ``0'' to ``9''. 
In the deep learning era (from 2012), ImageNet \cite{imagenet-cvpr-2009} is the most popular database.
Its first version that was used for ILSVRC 2012 includes 1.28M images and 1000 object categories. 
And now it has been expanded to 14M images. 
Many deep learning models (\eg, AlexNet \cite{alexnet-cnn-nips-2012}, VGG \cite{vgg-f-bmvc-2014} and ResNet \cite{residual-he-cvpr-2016}) are trained and evaluated on ImageNet to demonstrate their performance, and the checkpoints pre-trained on ImageNet are widely used to help other vision tasks, such as image annotation, object detection, etc. 
WebVision \cite{webvision-2017} includes 2.4M images, with the same 1,000 object categories in ImageNet.
The main difference between WebVision and ImageNet is that the annotations of WebVision are noisy, while the annotations of ImageNet are accurate. 
However, the authors of WebVision have experimentally demonstrated that the AlexNet trained on sufficient images with noisy labels has comparable or even better performance in visual representation than those trained on ImageNet. 
In addition to above image databases of object categories, there are two popular databases of scene categories, including SUN and Places. 
SUN \cite{sun-cvpr-2010} includes 108,754 images, with 397 scene semantic categories. 
Places \cite{places-2017} includes 10M images, with 434 scene semantic categories. 
However, scene categories describe higher level information than object categories. The visual representation of the deep model trained on scene databases may not be suitable for other vision tasks like object recognition or detection. 
However, as mentioned in section \ref{sec: introduction}, the main contents of one image cannot be well described by a single label. Visual representation learning on single-label images will waste useful information of training images, and may bring in confusion to deep models, as two visually similar images could be annotated with two different categories. 

There are also many multi-label image databases. 
Before the deep learning era, most multi-label image databases are used to evaluate multi-label models or image annotation methods. Some widely used databases include 
Corel 5k \cite{corel5k-eccv-2002} (including 4,999 images with 260 object categories), 
ESP Game \cite{espgame-2004} (including 20,770 images with 268 categories), 
IAPRTC-12 \cite{iaprtc-12-data-2006} (including 19,627 images with 291 categories),
NUSWIDE \cite{nus-wide-civr09} (including 270K images with 81 categories), 
MS COCO \cite{mscoco-2014} (including 330K images with 80 categories), and 
PASCAL VOC 2007 \cite{pascal-voc-2007} (including 9,963 images with 2.47 averaged annotated tags per image).
However, they are rarely used to train deep models for visual representation learning. Their scales are not big enough to train good parameters of popular deep models, such as VGG or ResNet. Besides, their small-scale category vocabularies are not diverse enough to train models with good generalization to other vision tasks. 
In contrast, there are also large-scale multi-label image databases. 
For example, Open Images-v1 \cite{openimages} includes 9M image with 6K categories. 
JFT-300M is an ``internal dataset'' in Google, including 300M images and 18,291 categories, as well as 1.25 averaged annotated tags per image.   
Sun et al. \cite{google-JFT300m-iccv-2017} trained the ResNet-101 model on JFT-300M, and transferred the trained checkpoint to other vision tasks, including single-label image classification on ImageNet, object detection on MS-COCO and PASCAL VOC 2007, semantic segmentation on PASCAL VOC 2012, and human pose estimation on MS-COCO.
Specifically, the checkpoint of ResNet-101 pre-trained on JFT-300M is fine-tuned on ImageNet, leading to $79.2\%$ top-1 accuracy on the validation set of ImageNet. In contrast, the ResNet-101 model trained on ImageNet from scratch gives $77.5\%$ top-1 accuracy. 
This improvement demonstrates that JFT-300M is helpful for learning more generalized visual representation. 
However, it is notable that the scale of JFT-300M is more than 200 times of ImageNet. Training ResNet-101 on 300M images with 18,291 categories is very costly. As reported in \cite{google-JFT300m-iccv-2017}, their training process takes 2 months for 4 epochs, using ``asynchronous gradient descent training on 50 NVIDIA K80 GPUs and 17 parameter servers". 
Moreover, JFT-300M and its checkpoints have not been published. They cannot be utilized by the research community to help other vision tasks. 
In contrast, the built Tencent ML-Images is publicly available, and our training based on distributed training framework is much more efficient. 

\section{The Tencent ML-Images Database}
\label{sec: dataset}

\subsection{Image Source and Class Vocabulary}
\label{sec: subsec image source and vocabulary}

The images and class vocabulary of Tencent ML-Images are collected from ImageNet \cite{imagenet-cvpr-2009} and Open Images \cite{openimages}. 
In the following we introduce the construction of training set, validation set and class vocabulary, respectively. 

Firstly, we extract image URLs from ImageNet-11k\footnote{Downloaded from http://data.mxnet.io/models/imagenet-11k/}. 
It is a subset of the whole database of ImageNet, collected by MXNet. 
It originally includes 11,797,630 training images, covering 11,221 categories. 
However, 1,989 categories out of 11,221 categories are very abstract in visual domain, such as {\it event} and {\it summer}. We think the images annotated with such abstract categories is helpless for visual representation learning. Thus, we remove these abstract categories and their corresponding images, with 10,322,935 images of 9,232 categories left.  
Moreover, according to the semantic relationship among categories, we add other 800 finer-grained categories from the whole vocabulary of ImageNet. For example, if {\it dog} is included in the above 9,232 categories, we also add {\it Husky} into the vocabulary of Tencent ML-Images, as well as the corresponding images from ImageNet.  
Consequently, we obtain 10,756,941 images, covering 10,032 categories from ImageNet.
We randomly select 50,000 images as validation set, while ensuring that the number of selected images of each category is no larger than 5. 
On the other hand, the Open Images-v1 contains about 9M images and 6K categories. We filter all images of Open Images using a per-category criteria. If one category occurs in less than 650 images, then we remove this category. 
We also remove some abstract categories in visual domain as did above. 
Besides, as some categories from Open Images are similar to or synonyms of the above 10,032 categories, we merge these redundant categories into unique ones. 
If all tags of one image are removed, then this image is also abandoned. 
Consequently, 6,902,811 training images and 38,739 validation images are remained, covering 1,134 unique categories. 
Finally, we merge the selected images and categories from ImageNet and Open Images to construct the Tencent ML-Images database, which includes 17,609,752 training and 88,739 validation images, covering 11,166 categories.

\subsection{Tag Augmentation of Images} 
\label{sec: tag augmentation}

Note that each image from ImageNet-11K is originally annotated by a single tag. As analyzed above, there usually exists multiple contents in one nature image. Single tag annotation may not cover the whole content in each image, and thus misses some helpful information. Meanwhile, as the size of class vocabulary is very large, there may also exist missing tags for multi-label annotations in Open Images. Due to the large scale of images and class vocabulary, it is challenging and time-consuming to manually augment the tags for each image.
We thus propose to augment the tags of these images by utilizing the semantic hierarchy and the co-occurrence among categories as follows. 

We firstly map the categories of Tencent ML-Images to the WordIDs in WordNet. According to the WordIDs, we construct the semantic hierarchy among these 11,166 categories.  It include 4 independent trees, of which the root nodes are {\it thing, matter, physical object} and {\it atmospheric phenomenon}, respectively. The length of the longest semantic path from root to leaf nodes is 16, and the  average length is $7.47$. The constructed semantic hierarchy captures the semantic relations among different categories and is used for tag augmentation.
Specifically, according to the semantic hierarchy, all ancestor categories of the original tag are also annotated as positive tags of the same image. \yanbo{For example, if one image is originally annotated as ``dog'', we also label is as ``animal''.}
Secondly, we compute the co-occurrence matrix $CO$ between categories from ImageNet-11k and categories from Open Images. Specifically, we train a ResNet-101 model with 1,134 outputs, based on Open Images. Using this trained model, we predict the labels among these 1,134 categories for the images from ImageNet-11k. 
If the posterior probability with respect to one category of one image is larger than $0.95$, then we set this category as the positive tag of this image. 
Then, we compute the co-occurrence matrix as follows: for category $i$ from ImageNet-11k and category $j$ from Open Images, we denote the number of positive images of category $i$ in ImageNet-11k as $n_i$, among which the number of images also annotated as category $j$ is denoted as $n_{i,j}$, then $CO(i,j) = n_{i,j} / n_i \in [0,1]$.  
If $CO(i,j) > 0.5$ and there is no semantic relationship between category $i$ and $j$ (\ie, there is no path from $i$ to $j$ or reverse in the semantic hierarchy), then we determine that category $i$ and $j$ is a strongly co-occurrent pair of categories. 
Then, we augment the tags of images from ImageNet-11k as follows: if one image is originally annotated as $i$, then we also label it as category $j$. \yanbo{For example, if an image is originally annotated as “sea snake”, then we also label it as “sea”.}

\subsection{Data Statistics} 
\label{sec:data-statistic}

\begin{figure}[t]
	\centering
	\includegraphics[width=0.49\textwidth,height=1.7in]{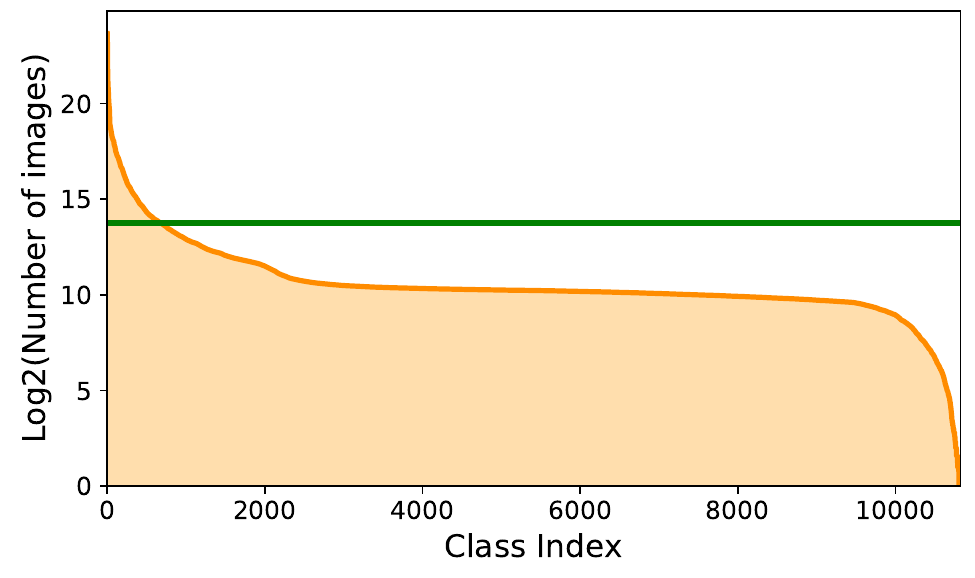}
	\vspace{-2em}
	\caption{Number of images ($\log_{2}$) per category in Tencent ML-Images. The \yanbo{green line} indicates the average number of images of all categories.}
	\label{fig-num-images-per-category}
	\vspace{-1em}
\end{figure}

\begin{figure}[t]
	\centering
	\includegraphics[width=0.49\textwidth,height=1.9in]{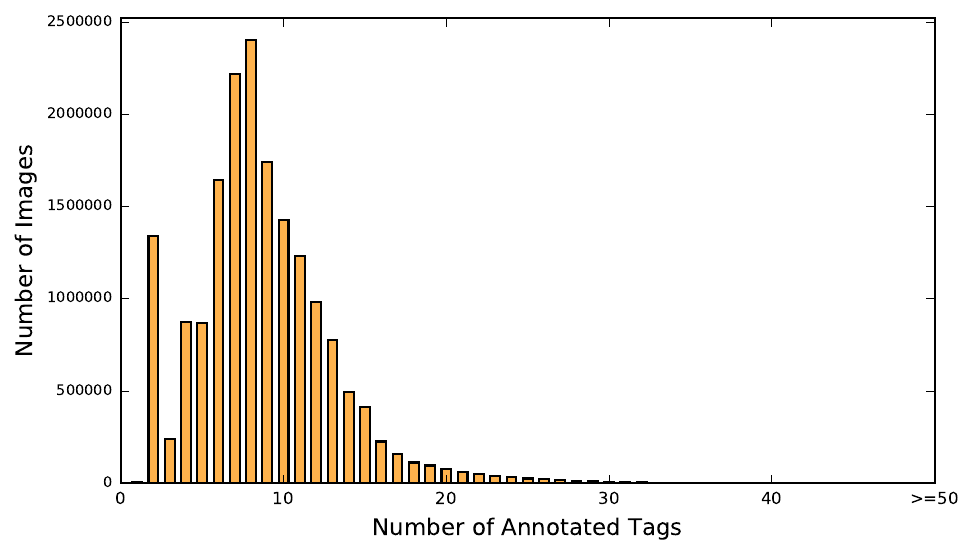}
	\vspace{-2em}
	\caption{The statistics of the numbers of annotated tags of all training images in Tencent ML-Images.}
	\label{fig-num-tags-all-images}
	\vspace{-1em}
\end{figure}

\vspace{2pt}
\noindent
{\bf Distribution of annotations}.
The number of images per category is shown in Fig. \ref{fig-num-images-per-category}. Specifically, the maximum number of images per category is 13,217,523, corresponding to the category ``object, physical object''; the minimum number is 0; the average number is 13,843. 
The distributions of different categories are extremely imbalanced. Some categories are frequent, while many others are very rare. 
It is referred to as the imbalance among categories \cite{my-aaai-2016-imbalance}.
There are 10,505 trainable categories, of which the numbers of images are larger than 100. 
The statistics of the numbers of annotated tags of all training images are shown in Fig. \ref{fig-num-tags-all-images}. 
Specifically, the numbers of annotated tags of all training images range from $1$ to $91$, and the average number is $8.72$. 
Considering the size of the label vocabulary (\ie, 11K), the number of annotated tags per image is very small. In other words, the number of positive tags of each image is much smaller than the number of negative tags. %
It is referred to as the imbalance between positive and negative tags per image \cite{my-aaai-2016-imbalance}. 
Above two types of imbalance bring in difficulty to model training. They will be considered during the training process of our model, as shown in section \ref{sec: experiment multi-label}. 

\vspace{2pt}
\noindent
{\bf Noisy and missing tags}. 
Noisy tag indicates the incorrectly annotated tag. \yanbo{And} missing tag \cite{my-icpr-2014,my-iccv-2015,wu2015multi,li2016facial,my-ijcv-2018} means that one class occurs in the image, but it is not annotated. 
As demonstrated in \cite{openimages}, the annotated tags for most images in Open Images are generated by machine, while the annotations of only a few fraction of images are verified by humans. \yanbo{The noisy annotations are unavoidable and they are also included in Tencent ML-Images.}
%
\yanbo{Most missing tags occur in images from ImageNet-11K, as they are originally annotated by single tag.}
As demonstrated in section \ref{sec: tag augmentation}, 
we augment the tags of these single-label \yanbo{annotations} by the category co-occurrence and semantic hierarchy. Compared to automatically generating tags by machine as did in Open Images, our augmentation is rather conservative. 
Our concern is that it is difficult to control the noise proportion of the machine-generated annotations, and we believe that the negative influence of noisy tags is larger than that of missing ones. 
Some works (\eg, \cite{openimages-noisy-learning-cvpr-2017,google-JFT300m-iccv-2017}) have demonstrated that learning from massive noisy labeled images is still able to show good visual representation. But they have not studied the trade-off between noisy and missing annotations, as the accurate proportions of both types of annotations are costly to calculate on large-scale databases. %
In this work, we choose the setting of more missing but less noisy annotations.

\section{Visual Representation Learning on Tencent ML-Images}
\label{sec: experiment multi-label}

\subsection{Training ResNet-101 with Multi-label Outputs on Tencent ML-Images}
\label{sec: experiment multi-label resnet-101}

\subsubsection{Model and Loss Function}
For visual representation learning on Tencent ML-Images databse, we implement the popular ResNet-101 model. As our task is multi-label classification, the outputs of ResNet-101 are the activations of $m$ independent Sigmoid functions, with $m$ being the size of the class vocabulary.
To alleviate the imbalance problems described in section \ref{sec:data-statistic}, we propose a novel weighted cross entropy loss function. For clarity, in the following, we present the loss function with respect to one training image $\x_i$:
\begin{flalign}
\mathcal{L}_{\boldsymbol{W}}(\x_i, \y_i) = \frac{1}{m} \sum_j^m r_t^{y_{ij}} \bigg[ - \eta y_{ij} \log(p_{ij}) - (1-y_{ij})\log(1-p_{ij}) \bigg],
\label{eq: loss function}
\end{flalign}
where $p_{ij} = f_{\W}(\x_i, j) \in [0,1]$ denotes the posterior probability with respect to category $j$, with $\W$ being the trainable parameters.
$\y_i = [y_{i1}, \ldots, y_{ij}, \ldots, y_{im}] \in \{0,1\}^m$ indicates the ground-truth label vector of image $\x_i$. 

\begin{itemize}
	\item \yanbo{Due to the highly imbalance of Tencent ML-Images, for many categories, the number of positive images is far less than that of negative ones}. The cost parameter $\eta>1$ is thus introduced to set a larger cost on positive labels than negative labels, to alleviate the imbalance between positive and negative images in each category, \yanbo{which is a common strategy in imbalance learning \cite{he2009learning}}. In our experiments, $\eta$ is set to 12. 
	\item $r_t^{y_{ij}}$ denotes an adaptive weight during the training process. It is formulated as follows:
	\begin{flalign}
	r_t^{y_{ij}} = 
	\begin{cases}
	\max\{0.01, \log_{10}(\frac{10}{0.01+t})\} \in [0.01, 1), & \text{if} y_{ij}=1; \\
	\max\{0.01, \log_{10}(\frac{10}{8+t})\} \in [0.01, 0.1), & \text{if} y_{ij}=0.
	\end{cases}
	\end{flalign}
	For category $j$, if all training images in one mini-batch are negative, then we record the status as $0$; if at least one training image in this mini-batch is positive, then we record the status as $1$. Consequently, we record a status vector like ($\ldots, 0, 1, 1, 1, 0, 0, 1, 0, \ldots$). Then, $t$ is defined as follows: if the status of the current mini-batch is different with that of the previous mini-batch, \ie, 01 or 10, then $t=1$; if the current status is same with the previous status, then $t = t+1$. With $r_t^{y_{ij}}$, if the parameters corresponding to category $j$ are positively or negatively updated in sequential mini-batches, the weight of the corresponding loss is decayed. It helps to alleviate the imbalance between frequent and rare categories. Besides, as the positive sequential mini-batches is less frequent than the negative sequential mini-batches, we set $r_t^{1} > r_t^{0}$ to alleviate the imbalance between positive and negative labels. 
\end{itemize}

\subsubsection{Image Pre-processing}
\label{sec: preprocessing}
\yanbo{For image pre-processing, data augmentation and normalization are widely adopted in image classification to improve the generalization \cite{alexnet-cnn-nips-2012,szegedy2015going,residual-he-cvpr-2016}. Following \cite{szegedy2015going}, our image pre-processing consists of the following six sequential steps. }
\begin{enumerate}
	\item Random crop a bounding box from input image, for which the box area is within $[0.05, 1.0]$ of the whole image area, and the aspect ratio between its width and height is within $[\frac{3}{4}, \frac{4}{3}]$.
	\item Resize the cropped box to $224 \times 224$.
	\item Random flip the cropped box horizontally with probability of $0.5$. 
	\item Random rotate the cropped box with probability of $0.25$, and the rotation degree is evenly sampled from $[-45, 45]$.
	\item Random shift the color with probability of $0.5$. 
	\item Linearly rescale pixel value to $[-1,1]$. 
\end{enumerate}
\yanbo{Besides, we use a relatively small value (\ie, 0.05) for the lower bound of box area ratio in step-1 to include more small patches. We also add a random rotation (\ie, step-4) for data augmentation. The range of rotation degree is set to [-45,45] experimentally.}

\begin{figure}[t]
	\centering
	\includegraphics[width=0.48\textwidth,height=2.4in]{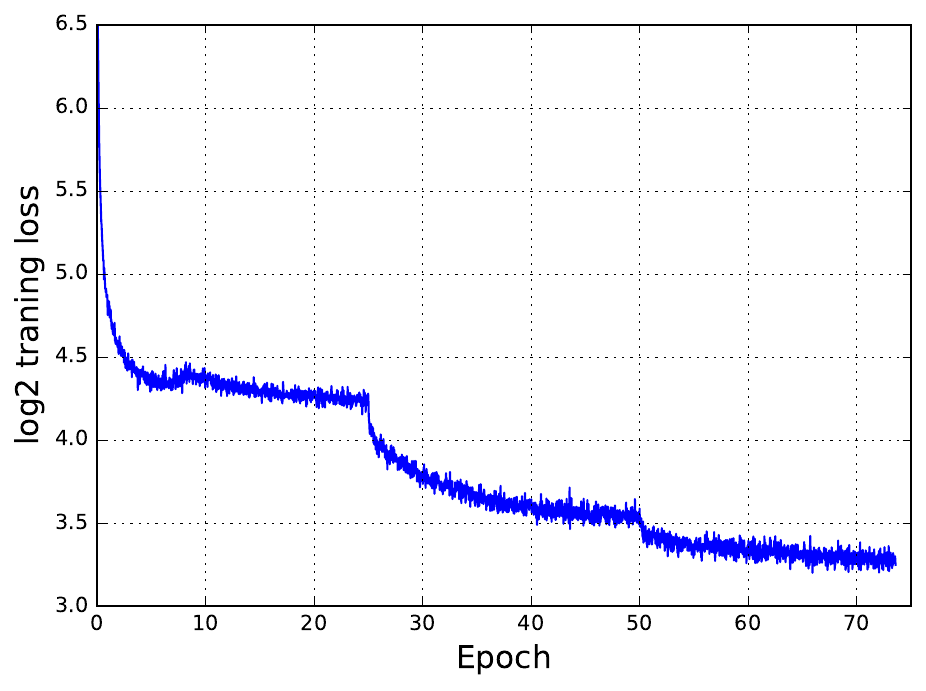}
	\vspace{-1em}
	\caption{The curve of the $\log_2$ training loss of pre-training the ResNet-101 model on Tencent ML-Images.}
	\label{fig-train-loss-curve}
	\vspace{-1em}
\end{figure}

\subsubsection{Training Algorithm and Hyper-parameters}
We adopt stochastic gradient descent (SGD) with momentum and back-propagation \cite{back-propagation-hinton-1986} to train the ResNet-101 model. 
%
There are 17,609,752 training images. \yanbo{To speed up the training process, we follow the ``large minibatch SGD'' and ``linear scaling rule'' proposed by \cite{warmup-2017}. Specifically, our batch-size is set to 4,096, and each epoch includes 4,300 steps. According to the ``linear scaling rule'', when the batch-size is multiplied by {\it k}, the learning rate should be multiplied by {\it k}. In our implementation, we adopt a reference learning rate of 0.01 for batch-size 512, and the learning rate is 0.08 for our large batch-size 4,096. Moreover, as the “linear scaling rule” may not hold in early epochs when the network is changing rapidly, warm-up strategy is proposed by \cite{warmup-2017}. The key idea of warm-up is to use less aggressive learning rates at the start of training. We adopt the gradual warm-up strategy in \cite{warmup-2017} that gradually ramps up the learning from a small (which is 0.01 in our implementation) to a desired value (which is 0.08 in our implementation). Our warm-up phase contains 8 epochs, and the learning rate increasing factor during warm-up is 1.297. After the warm-up phase, the learning rate reaches 0.08, and we go back to original learning rate decay schedule. That is, the learning rate decays with factor $0.1$ in every 25 epochs. The maximal epoch number is 60.}
The momentum is 0.9. For the updating of the parameters of BatchNorm, the decay factor of moving average is 0.9, and the constant $\epsilon$ is set to $0.001$ to avoid the $0$ value of the variance. The weight decay is $0.0001$. 

In each batch, for many categories, due to the highly imbalance between positive and negative images, most training images are negative. For category $i$, if there are no positive images in the current batch, the model parameters corresponding to category i will only be updated with a probability of $0.1$; if there are positive images for category $i$, we will down-sample negative images with the number of 5 times of positive images, then the model parameters are updated according to positive and down-sampled negative images. 
Besides, we set the cost parameter $\eta$ to 12, \ie, a higher cost for positive labels than negative ones. It helps to alleviate the negative influence of the imbalance between positive and negative images. The curve of the $\log_2$ training loss is shown in Fig. \ref{fig-train-loss-curve}.

\begin{figure}[t]
	\centering
	\includegraphics[width=0.49\textwidth,height=1.8in]{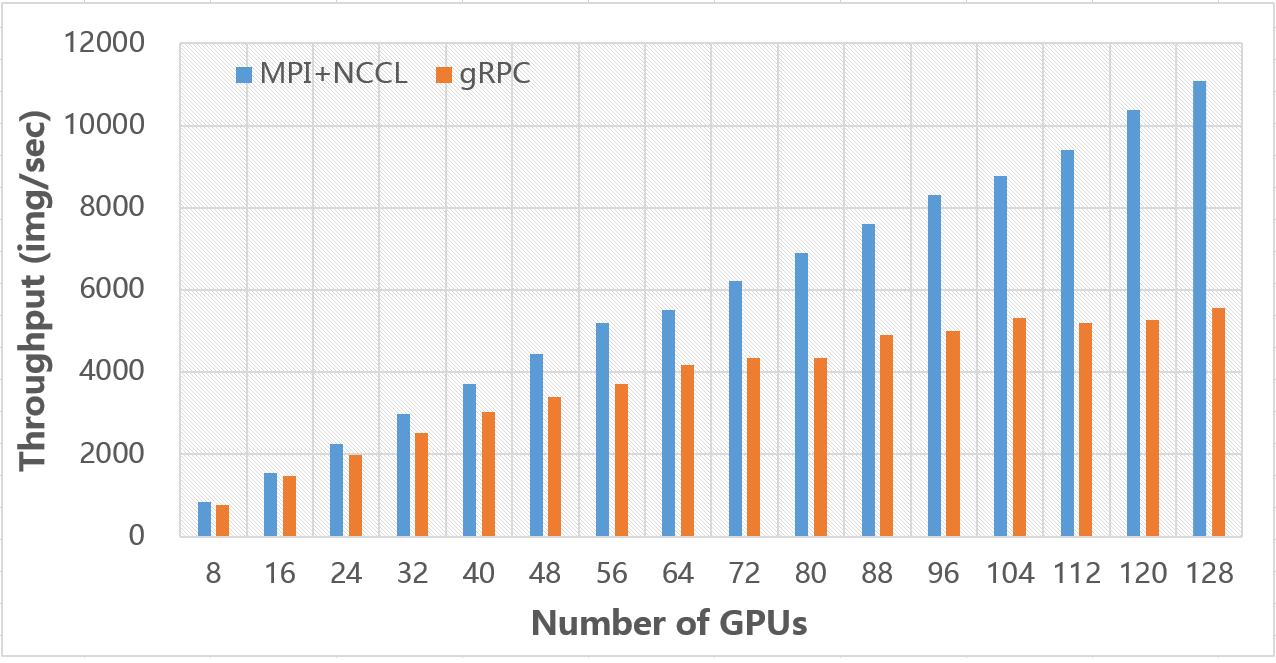}
	\vspace{-2em}
	\caption{Throughout of distributed training based on MPI+NCCL and gRPC.}
	\label{fig-speed}
	\vspace{-1em}
\end{figure}

\begin{figure}[t]
	\centering
	\includegraphics[width=0.49\textwidth,height=1.8in]{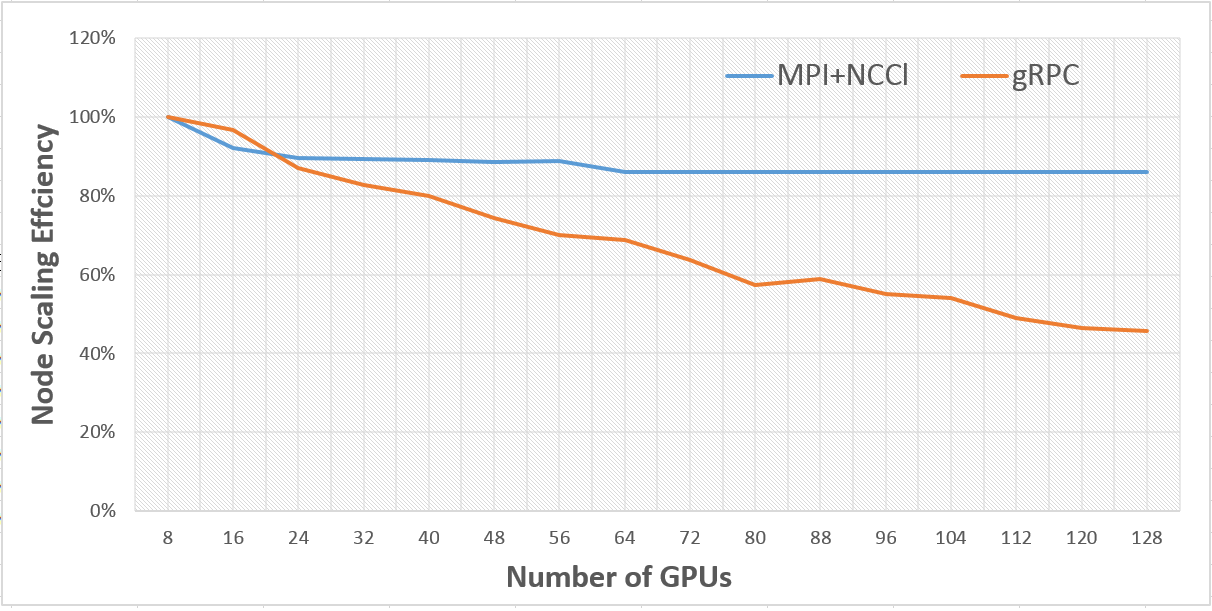}
	\vspace{-2em}
	\caption{Scaling efficiency of each GPU in distributed training based on MPI+NCCL and gRPC.}
	\label{fig-scaling-efficiency}
	\vspace{-1em}
\end{figure}

\subsubsection{Acceleration by Distributed Training}
The training of Resnet-101 model with 11K categories and 18M images requires lots of computation. It will take a few dozen days if using one single GPU. In this work, all training experiments are conducted on a large-scale distributed deep learning framework, \ie, TFplus, which is built upon Tensorflow with several communication optimized techniques. We replace the original gRPC implementation with Message Passing Interface (MPI) and NVIDIA Collective Communications Library (NCCL) \cite{NCCL-2017}. NCCL provides a highly optimized version of routines, such as all-gather, all-reduce, broadcast, reduce, reduce-scatter, and the integrated bandwidth-optimal ring all-reduce algorithm \cite{all-reduce-2009}, to achieve high bandwidth over PCIe on NVIDIA GPU. In order to scale from one GPU to multiple nodes and multiple GPUs, we implement several APIs for communication: 1) a broadcast operation to synchronize parameters among all GPUs at the initialization stage or the recovery from the checkpoint; 2) a distributed optimizer wrapper for synchronization update of parameters; 3) some operations for data partition and barrier, etc.  
Since both MPI and NCCL support the remote direct memory access (RDMA), we run all distributed training jobs over a 40-GbE RDMA-capable networking. 
We achieve about 2X speed up compared with the original gRPC based distributed implementation on a cluster of 16 nodes and each node with 8 NVIDIA M40 GPUs, as shown in Fig. \ref{fig-speed}. Specifically, when training with 128 GPUs, the throughout (\ie, the number of processed images per second) of MPI+NCCL is up to 11077, while the throughout of gRPC is 5551. 
Besides, the distributed training jobs based on MPI+NCCL achieve $~86\%$ scaling efficiency from 8 to 128 GPUs, while those based on gRPC are only $46\%$, as shown in Fig. \ref{fig-scaling-efficiency}.
The whole training process on Tencent ML-Images takes 90 hours for 60 epochs with 128 GPUs, \ie, 1.5 hours per epoch. 

\subsection{Evaluations and Results} 
\label{sec: experiment multi-label evaluation}

To evaluate the performance of the trained ResNet-101 model with multi-label outputs, we adopt the widely used instance-level metrics in multi-label learning, including instance-level precision, recall and F1 score \cite{zhang2013review}. 
As the output for each category is the posterior probability, we need to transform the continuous predictions to binary predictions to calculate above metrics. 
Specifically, for image $i$, we determine the categories corresponding to top-$k$ largest posterior probabilities as positive labels (\ie, $1$), while all other categories are negative labels (\ie, $0$). We obtain a binary prediction vector $\hat{\y}_i^k \in \{0,1\}^m$. 
Then, the evaluation metrics are calculated as follows \cite{zhang2013review}:
\begin{flalign}
P_k & = \frac{1}{n} \sum_i^n P_{i,k} = \frac{1}{n} \sum_i^n \frac{\y_i * \hat{\y}_i^k}{k}, 
\\
R_k & = \frac{1}{n} \sum_i^n R_{i,k} = \frac{1}{n} \sum_i^n \frac{\y_i * \hat{\y}_i^k}{\boldsymbol{1}^\top * \y_i}, 
\\
F1_k & = \frac{1}{n} \sum_i^n \frac{2 P_{i,k} * R_{i,k}}{P_{i,k} + R_{i,k}}.
\end{flalign}

\begin{table}[t] 
	\begin{center}
		\scalebox{1.05}{
			\begin{tabular}{p{.15\textwidth}| p{.07\textwidth} p{.07\textwidth} p{.07\textwidth} }
				\hline
				\scalebox{1}{Evaluation metric} $\rightarrow$ & \multirow{2}{*}{Precision} & \multirow{2}{*}{Recall} & \multirow{2}{*}{F1}
				\\
				\scalebox{1}{Top-k prediction} $\downarrow$ & & &  
				\\
				\hline \hline
				\scalebox{1}{Top-5 prediction} & 35.7 & 18.2 & 23.3
				\\
				\scalebox{1}{Top-10 prediction} & 29.0 & 29.5 & 28.1
				\\
				\hline
			\end{tabular}
		}
	\end{center}
	\caption{ Results ($\%$) of the ResNet-101 with multi-label outputs, evaluated on the validation set of Tencent ML-Images.}
	\label{table: results of multi-label classification}
	\vspace{-0.1in}
\end{table}

We present results of top-5 and top-10 predictions in Table \ref{table: results of multi-label classification}. 
The evaluation values are not very high. As demonstrated in section \ref{sec: subsec image source and vocabulary}, the size of the validation set is only about $\frac{1}{200}$ of the size of the training set. And there should be many missing labels for validation images. Thus, the evaluation scores on this small validation set are not reliable enough to measure the visual representation performance of the model trained on Tencent ML-Images. 
Instead, its performance could be evaluated through transfer learning to some other visual tasks, as follows. 

\section{Transfer Learning}
\label{sec: transfer learning}

\subsection{Transfer Learning to Single-Label Image Classification on ImageNet}
\label{sec: transfer learning to imagenet}

To verify the quality of the visual representation of the ResNet-101 model pre-trained on Tencent ML-Images, we conduct transfer learning to image classification on the benchmark single-label image database, \ie, ImageNet. Specifically, we utilize the ResNet-101 model pre-trained on Tencent ML-Images as the initial checkpoint, and replace the output layer with 1,000 output nodes, as well as the loss with standard softmax loss. Then, we fine-tune the model on ImageNet. Experimental results are given below.

\subsubsection{Fine-Tuning Approaches}
\label{sec: fine-tuning}

\noindent 
{\bf Learning rate}.
Note that there are significant differences between Tencent ML-Images and ImageNet. First, the distributions of visual features and class vocabulary are different. Second, images in Tencent ML-Images are annotated with multiple tags, while images in ImageNet are annotated with single label. Last, the annotations of Tencent ML-Images are noisy, while the annotations of ImageNet are clean. 
Considering these significant differences, one cannot expect that the ResNet-101 model pre-trained on Tencent ML-Images show good classification performance on ImageNet, if without fine-tuning of the parameters. 
The standard fine-tuning approach adopts one consistent learning rate of all layers. It is referred to as fine-tuning with layer-wise consistent learning rates. 
However, as verified in later experiments, the ResNet-101 model with the above fine-tuning approach even shows worse performance than the ImageNet baseline. It reveals that the useful information contained in the Tencent ML-Images checkpoint has not been well utilized, due to the aforementioned three differences. 
To tackle this issue, we adopt the fine-tuning approach with layer-wise adaptive learning rates. Specifically, we set larger learning rates on top layers, while smaller learning rates on bottom layers. The rationale behind this setting is that the parameters of top-layers are more dependent on training images and labels, while the parameters in bottom layers represent low-level visual features. To alleviate the negative influence of the significant differences between Tencent ML-Images and ImageNet, the top-layers' parameters should be changed to be further from the initialized parameters from the checkpoint, compared to the bottom layers' parameters. 

\noindent 
{\bf Image size}. 
As demonstrated in \cite{resnet-v2-eccv-2016}, the image size in training and testing has a significant influence to the results. Further, YOLO9000 \cite{yolo9000} proposes to adjust the image size during training, which is proved to be helpful for the detection performance. 
Besides, we note that the image size of pre-training, fine-tuning and evaluation used in  \cite{google-JFT300m-iccv-2017} is $299 \times 299$, while the image size in our pre-training is $224 \times 224$. 
To conduct a fair comparison with the JFT-300M, and also to explore the influence of image size to image classification, we design three fine-tuning settings with different image sizes, including: 1) the image size is kept at $224\times 224$; 2) the image size in early epochs is set as $224\times 224$, while $299\times 299$ in late epochs; 3) the image size is kept at $299\times 299$. 

\begin{table}[t] 
	\begin{center}
		\scalebox{1.06}{
			\begin{tabular}{p{.15\textwidth}| p{.07\textwidth} p{.07\textwidth} p{.07\textwidth} }
				\hline
				Hyper-parameters & \scalebox{0.8}{ckpt-1} & \scalebox{0.8}{ckpt-2} & \scalebox{0.8}{ckpt-3/4/5}
				\\
				\hline 
				Batch size & \multicolumn{3}{c}{2048} 
				\\
				Maximum epoch & \multicolumn{3}{c}{120}
				\\
				LR-top2-stages & 0.8 & 0.008 & 0.8
				\\
				LR-bottom2-stages & 0.8 & 0.008 & 0
				\\
				LR-decay-factor &  \multicolumn{3}{c}{0.1}
				\\
				LR-decay-step &  \multicolumn{3}{c}{18750 (30 epochs)}
				\\
				Weight-decay & \multicolumn{3}{c}{0.0001}
				\\
				warmup-steps & \multicolumn{3}{c}{2500 (the first 4 epochs)}
				\\
				LR-warmup & 0.1 & 0.001 & 0.1
				\\
				\scalebox{0.8}{LR-warmup-decay-factor} & \multicolumn{3}{c}{1.681}
				\\
				\scalebox{0.8}{LR-warmup-decay-step} & \multicolumn{3}{c}{625 (one epoch)}
				\\
				BatchNorm-decay & \multicolumn{3}{c}{0.9}
				\\
				BatchNorm-eps &  \multicolumn{3}{c}{0.001}
				\\
				\hline
			\end{tabular}
		}
	\end{center}
	\caption{ \small Hyper-parameters of different checkpoints in transfer learning to ImageNet.
		``ckpt'' denotes our ResNet-101 checkpoint. ckpt-1 means the checkpoint trained on ImageNet from scratch; ckpt-2 indicates the checkpoint pre-trained on Tencent ML-Images and fine-tuned on ImageNet with layer-wise consistent learning rates; ckpt-3/4/5 represent three checkpoints pre-trained on Tencent ML-Images and fine-tuned on ImageNet with layer-wise adaptive learning rates, and they are different at the image size in fine-tuning. 
		``LR'' denotes learning rate. ResNet-101 consists of 4 stages of residual blocks, and top2-stages indicates the two stages close to the output layer, while bottom2-stages represents those close to the input layer. Note that if there is only one value in a row, then it means that the checkpoints of different columns are trained with the same value of the corresponding hyper-parameter.}
	\label{table: hyper-parameters of fine-tuning}
	\vspace{-0.15in}
\end{table}

\subsubsection{Comparisons and Hyper-parameters}

To verify the quality of visual representation of the Tencent ML-Images checkpoint, we compare five ResNet-101 checkpoints with different fine-tuning approaches.  
The hyper-parameters of training these checkpoints are summarized in Table \ref{table: hyper-parameters of fine-tuning}.
Besides, we also present the reported results of others' implementations, including: the MSRA checkpoint of training on ImageNet from scratch; the Google's checkpoint of training on ImageNet from scratch \cite{google-JFT300m-iccv-2017}; the checkpoint of pre-training on JFT-300M and fine-tuning on ImageNet \cite{google-JFT300m-iccv-2017}. Please refer to Table \ref{table: results of ImageNet classification}.

\begin{table*}[t] 
	\begin{center}
		\scalebox{1.07}{
			\begin{tabular}{ p{.16\textwidth} |p{.4\textwidth}| p{.04\textwidth} p{.04\textwidth} | p{.04\textwidth} p{.04\textwidth} }
				\hline
				\multirow{2}{*}{\scalebox{1}{ResNet-101 checkpoints}} & \multirow{2}{*}{\scalebox{1}{Training and fine-tuning settings}}  & 
				\multicolumn{2}{c|}{Validation size 224} & \multicolumn{2}{c}{Validation size 299}
				\\
				&  &  \scalebox{0.9}{top-1 acc} & \scalebox{0.9}{top-5 acc} &  \scalebox{0.9}{top-1 acc} & \scalebox{0.9}{top-5 acc}
				\\
				\hline \hline
				\scalebox{0.9}{MSRA ckpt} \cite{residual-he-cvpr-2016} & Train on ImageNet, 224 & 76.4 & 92.9 & -- & --
				\\
				\scalebox{0.9}{Google ckpt-1} \cite{google-JFT300m-iccv-2017} & Train on ImageNet, 299 &  -- & -- & 77.5 &  93.9
				\\
				\scalebox{0.9}{Our ckpt-1} & Train on ImageNet, 224 &  77.8 & 93.9 & 79.0 & 94.5
				\\
				\hline \hline
				\scalebox{0.9}{Google ckpt-2}  \cite{google-JFT300m-iccv-2017} & Train on JFT-300M, fine-tune on ImageNet, 299 & -- & -- & 79.2 & 94.7
				\\
				\scalebox{0.9}{Our ckpt-2}  & Pre-train on Tencent ML-Images, fine-tune on ImageNet with layer-wise consistent learning rate & 74.7 & 92.6 & -- &  --
				\\
				\scalebox{0.9}{Our ckpt-3}  & Pre-train on Tencent ML-Images, fine-tune on ImageNet & {\bf 78.8} & {\bf 94.5} & 79.5 & 94.9
				\\
				\scalebox{0.9}{Our ckpt-4}  & Pre-train on Tencent ML-Images, fine-tune on ImageNet 224 to 299 & 78.3 & 94.2 & {\bf 80.73} & {\bf 95.5}
				\\
				\scalebox{0.9}{Our ckpt-5} & Pre-train on Tencent ML-Images, fine-tune on ImageNet 299 & 75.8 & 92.7 & 79.6 & 94.6
				\\
				\hline
			\end{tabular}
		}
	\end{center}
	\caption{ \small Results ($\%$) of single-label image classification on the validation set of ImageNet, using single model and single crop inference. 
		In the column of ``Train and fine-tune settings'', ``224'' indicates that the size of training image is $224 \times 224$, while ``299'' indicating $299 \times 299$; ``224 to 299'' means that the size of training image in early epochs is $224 \times 224$, and that in later epochs is $299 \times 299$. For Our ckpt-3/4/5, we adopt the layer-wise adaptive learning rate in fine-tuning.
	}
	\label{table: results of ImageNet classification}
	\vspace{-0.15in}
\end{table*}

\subsubsection{Results}

\begin{figure}[t]
	\centering
	\includegraphics[width=0.49\textwidth,height=2.5in]{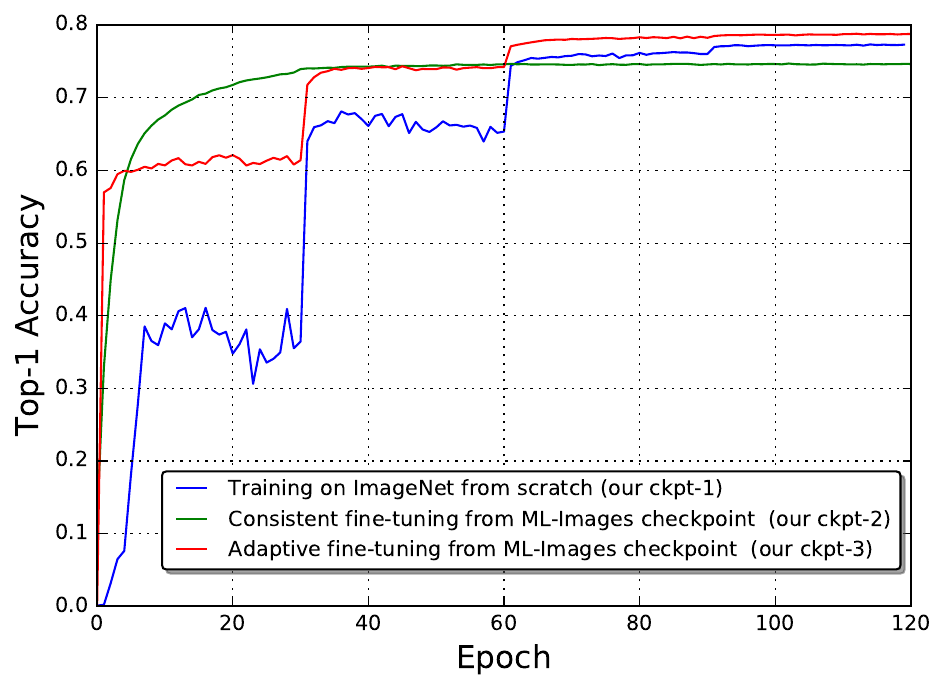}
	\vspace{-2em}
	\caption{The curves of top-1 accuracy of different checkpoints on the validation set of ImageNet.}
	\label{fig-top1-accuracy-imagenet}
	\vspace{-0.1in}
\end{figure}

All compared results are shown in Table \ref{table: results of ImageNet classification}. 
{\bf (1)} In terms of the baseline (\ie, train on ImageNet from scratch), our ckpt-1 is higher than both MSRA ckpt \cite{residual-he-cvpr-2016} and Google ckpt-1 \cite{google-JFT300m-iccv-2017}. 
Our implementation and MSRA implementation are same at the model architecture and the size of the input image  (\ie, $224 \times 224$). The main difference is the pre-processing of the input image (see section \ref{sec: preprocessing}) and hyper-parameters, which should be the main reasons of different performance of this two baselines. 
In contrast, the details of the model architecture and the image pre-processing are not demonstrated in \cite{google-JFT300m-iccv-2017}, and the image size is $299 \times 299$. But our ckpt-1 still performs better than Google ckpt-1 on the evaluation with the size of $299 \times 299$. 
It demonstrates the good quality of our implementation of the baseline.
{\bf (2)} 
Moreover, the comparison between our baseline checkpoint (\ie, our ckpt-1 in Table \ref{table: results of ImageNet classification}) and our fine-tuning checkpoints (\ie, our ckpt-2/3/4/5 in Table \ref{table: results of ImageNet classification}) demonstrate two points. %
The accuracy of our ckpt-2 is much lower than that of our ckpt-1. As analyzed in section \ref{sec: fine-tuning}, the significant difference between Tencent ML-Images and ImageNet, as well as the label noises in Tencent ML-Images, could bring in negative influence to the model performance. 
In contrast, our ckpt-3 with the layer-wise adaptive fine-tuning learning rate shows the improvements of $1.0\%$ at top-1 accuracy and $0.6\%$ at top-5 accuracy under validation size of $224 \times 224$. This demonstrates that the fine-tuning with layer-wise adaptive learning rate can not only utilize the good visual representation encoded in bottom layers of the Tencent ML-Images checkpoint, but also alleviate the significant difference between these two databases.
It reveals that the Tencent ML-Images checkpoint includes good visual representation, but it should be carefully fine-tuned to help other vision tasks. 
We also evaluate the checkpoints of different epochs using top-1 accuracy on the validation set of ImageNet, as shown in Fig.  \ref{fig-top1-accuracy-imagenet}.
{\bf (3)} 
The results of Google ckpt-2 are $79.2\%$ top-1 accuracy and $94.7\%$ top-5 accuracy. The improvements over Google ckpt-1 are $1.7\%$ at top-1 accuracy and $0.8\%$ at top-5 accuracy. It demonstrates the good quality of the initial checkpoint pre-trained on JFT-300M. 
In contrast, all of our ckpt-3/4/5 show higher top-1 accuracies than Google ckpt-2 when the image size of validation set is set to $299 \times 299$. And, our ckpt-4 with adaptive image sizes in fine-tuning achieves the highest $80.73\%$ top-1 and $95.5\%$ top-5 accuracy, and the improvements over our ckpt-1 are $1.73\%$ at top-1 and $1\%$ at top-5 accuracy. 
Our checkpoints exceed Google ckpt-2 on both the accuracy and the accuracy improvement over baseline. Considering that the size of JFT-300M is about 17 times of Tencent ML-Images, these verify the high quality of Tencent ML-Images and our training and fine-tuning. 

\subsection{Transfer Learning to Caltech-256 }
\label{sec: transfer learning to other datasets}

We also conduct transfer learning to another small-scale single-label image database, \ie, Caltech-256 \cite{caltech-256}, which includes 30,607 images with 256 object categories. 
We utilize a pre-trained checkpoint of the ResNet-101 model to extract features for each image of Caltech-256. Specifically, we adopt the output of the global  average pooling in ResNet-101 as the feature vector (2,048 dimensions). Then we train a multi-category SVM classifier to predict the labels for each image. 
We compare with three checkpoints, including: training on ImageNet from scratch; training on Tencent ML-Images from scratch; pre-training on Tencent ML-Images, and fine-tuning on ImageNet with layer-wise adaptive learning rate. 
The results of different checkpoints are shown in Table \ref{table: results of caltech-256}. The result of the ImageNet checkpoint is higher than that of the Tencent ML-Images checkpoint. It again demonstrates that the significant difference of distribution between Tencent ML-Images and single-label image databases. 
The checkpoint of adaptive fine-tuning from Tencent ML-Images gives $86.5\%$ accuracy, which is higher than $86\%$ accuracy of the ImageNet checkpoint. It also verifies that the good generalization of visual representation of the Tencent ML-Images could be well explored through adaptive fine-tuning.

\begin{table}[t] 
	\begin{center}
		\scalebox{0.98}{
			\begin{tabular}{p{.1\textwidth}| p{.1\textwidth} p{.1\textwidth} p{.1\textwidth} }
				\hline
				Checkpoints & Checkpoint 1 & Checkpoint 2 & Checkpoint 3
				\\
				\hline 
				Accuracy & 86.0 & 83.6 &  86.5
				\\
				\hline
			\end{tabular}
		}
	\end{center}
	\caption{ \small Classification accuracy ($\%$) of transfer learning to Caltech-256. Checkpoints 1 to 3 respectively indicate: training on ImageNet from scratch; training on Tencent ML-Images from scratch; pre-training on Tencent ML-Images, and fine-tuning on ImageNet with a layer-wise adaptive learning rate.}
	\label{table: results of caltech-256}
	\vspace{-0.1in}
\end{table}

\subsection{Transfer Learning to Object Detection} 
\label{sec: transfer learning to detection}

We conduct transfer learning to object detection on the benchmark PASCAL VOC database \cite{pascal-voc-2007,everingham2015pascal}, including 20 categories. To fairly compare with \cite{google-JFT300m-iccv-2017}, we also adopt the ``trainval''  images from both PASCAL VOC 2007 and 2012 as the training set, including 16,551 training images.
All models are evaluated on the testing set of PASCAL VOC 2007, including 4,952 images. We use mean average precision at $50\%$ IOU threshold (mAP@.5) for performance evaluation. 

{\bf Comparisons.} We compare with the transfer learning did in  \cite{google-JFT300m-iccv-2017}, including their baseline checkpoint pre-trained on ImageNet, and their checkpoints pre-trained on JFT-300M and JFT-300M+ImageNet. 
Our first checkpoint (\ie, our ckpt-1) is also pre-trained on ImageNet from scratch, then fine-tuned on VOC. 
Our second checkpoint (\ie, our ckpt-2) is firstly pre-trained on Tencent ML-Images, then fine-tuned on ImageNet (using the same setting with ``Our ckpt-3'' in Table \ref{table: results of ImageNet classification}), and further fine-tuned on VOC. 

{\bf Implementation details.} Our implementation is based on the TensorFlow implementation of the Faster RCNN framework \cite{cnn-object-detection-nips-2015,chen2017spatial}. Specifically, we use stochastic gradient descent with the momentum of $0.9$ for training; the initial learning rate is set to $8\times 10^{-4}$ and decays by $0.1$ at every $80$k steps; the batch size is set to $256$; the model is trained for $180$k steps; the weight decay is set to $10^{-3}$. The input image is resized such that the short side is fixed to $600$-pixels, while maintaining the aspect ratio.

{\bf Results.} 
The results are summarized in Table \ref{table: results of detection}.
In comparison of the baseline checkpoints, our ckpt-1 achieves $80.1\%$, while Google ckpt-1 gives $76.3\%$. 
With the same databases and models, it demonstrates that our implementation quality is much better than Google's implementation.  
Our ckpt-2 shows the improvement of $1.4\%$ than our ckpt-1. This verifies the good visual representation of the checkpoint pre-trained on Tencent ML-Images. 
In contrast, Google ckpt-2 and ckpt-3 show $81.4\%$ and $81.3\%$, respectively, which are slightly lower than our ckpt-2. 
It demonstrates that the checkpoint pre-trained on Tencent ML-Images and fine-tuned on ImageNet has similar generalization with those pre-trained on JFT-300M and JFT-300M+ImageNet. 
Considering that JFT-300M is about 17 times larger than our Tencent ML-Images, we could claim that Tencent ML-Images is a high-quality database. 
We also tried the checkpoint that is pre-trained on Tencent ML-Images, then fine-tuned on VOC. 
But the performance is not as good as other checkpoints, thus we didn't report its results here. We think the reason is that the big gap of the data distributions and tasks, between Tencent ML-Images and VOC. 

\begin{table}[t] 
	\begin{center}
		\scalebox{1}{
			\begin{tabular}{p{.077\textwidth}| p{.28\textwidth} p{.045\textwidth} }
				\hline
				Checkpoints & Pre-training and fine-tuning settings & \scalebox{0.8}{mAP@0.5}
				\\
				\hline \hline
				\scalebox{0.8}{Google ckpt-1} & Pre-train on ImageNet, fine-tune on VOC
				&  76.3
				\\
				\scalebox{0.8}{Our ckpt-1} & Pre-train on ImageNet, fine-tune on VOC & 80.1
				\\
				\hline
				\scalebox{0.8}{Google ckpt-2} & Pre-train on JFT-300M, fine-tune on VOC &  81.4
				\\
				\scalebox{0.8}{Google ckpt-3} &  \scalebox{0.8}{Pre-train on JFT-300M+ImageNet, fine-tune on VOC} & 81.3
				\\
				\multirow{2}{*}{\scalebox{0.8}{Our ckpt-2}} & Pre-train on Tencent ML-Images, fine-tune   & \multirow{2}{*}{\textbf{81.5}}
				\\
				& on ImageNet, and then fine-tune on VOC & 
				\\
				\hline
			\end{tabular}
		}
	\end{center}
	\caption{ \small Results of object detection on the testing set of PASCAL VOC 2007. Note that ``VOC'' in ``fine-tune on VOC'' indicates the combined training set of PASCAL VOC 2007 and 2012.}
	\label{table: results of detection}
	\vspace{-0.1in}
\end{table}

\subsection{Transfer Learning to Semantic Segmentation} 
\label{sec: transfer learning to segmentation}

We conduct transfer learning to semantic segmentation on the benchmark PASCAL VOC 2012 database \cite{everingham2015pascal}, including 20 foreground categories and 1 background category. To fairly compare with \cite{google-JFT300m-iccv-2017}, we also adopt the augmented training set of PASCAL VOC 2012 as the training set of fine-tuning, which includes 10,582 training images. All models are evaluated on the validation set of PASCAL VOC 2012, including 1,149 images, using the mean itersection-over-union (mIOU) metric. 

{\bf Comparisons.}  We adopt the same setting of checkpoints as did in the above object detection experiments (see section \ref{sec: transfer learning to detection}). Thus, here we didn't repeat it to keep clarity. 

{\bf Implementation details.} Our implementation is based on the semantic segmentation architecture of DeepLab \cite{chen2018deeplab}. For fair comparison, our implementation also adopts DeepLab-ASPP-L structure that has four branches after the Conv5 block of ResNet-101 model. All ASPP branchs use $3\times 3$ kernels but with different atrous rates (\ie, $\{6,12,18,24\}$). During training, we use the ``poly'' learning rate policy (with power = $0.9$) and the initial learning rate is set to $3 \times 10^{-3}$. The weight decay is set to $5 \times 10^{-4}$. The model is trained for $50$k steps using the stocastic gradient descent with momentum of $0.9$. The batch size is set to $6$, and the input image is resized to $513 \times 513$.

{\bf Results.} 
The results are summarized in Table \ref{table: results of segmentation}. 
We could obtain the similar observations with the transfer learning to object detection (see section \ref{sec: transfer learning to detection} and Table \ref{table: results of detection}). 
1) Our implementation of the baseline checkpoint is better than that of Google, \ie, our ckpt-1 $74.0\%$ vs. Google ckpt-1 $73.6\%$.
2) Our ckpt-2 shows the improvement up to $2.3\%$ over our ckpt-1, which verifies the good visual representation of the checkpoint pre-trained on Tencent ML-Images. 
3) Google ckpt-2 shows $75.3\%$ and Google ckpt-3 gives $76.5\%$. 
It demonstrates that the checkpoint pre-trained on Tencent ML-Images and fine-tuned on ImageNet has better generalization than that pre-trained on JFT-300M, while is similar with the checkpoint pre-trained on JFT-300M+ImageNet. 

\begin{table}[t] 
	\begin{center}
		\scalebox{1}{
			\begin{tabular}{p{.075\textwidth}| p{.28\textwidth} p{.045\textwidth} }
				\hline
				Checkpoints & Pre-training and fine-tuning settings & \scalebox{0.8}{mIOU}
				\\
				\hline \hline
				\scalebox{0.8}{Google ckpt-1} & Pre-train on ImageNet, fine-tune on VOC
				&  73.6
				\\
				\scalebox{0.8}{Our ckpt-1} & Pre-train on ImageNet, fine-tune on VOC & 74.0
				\\
				\hline
				\scalebox{0.8}{Google ckpt-2} & Pre-train on JFT-300M, fine-tune on VOC &  75.3
				\\
				\scalebox{0.8}{Google ckpt-3} &  \scalebox{0.8}{Pre-train on JFT-300M+ImageNet, fine-tune on VOC} & \textbf{76.5}
				\\
				\multirow{2}{*}{\scalebox{0.8}{Our ckpt-2}} & Pre-train on Tencent ML-Images, fine-tune  & \multirow{2}{*}{76.3}
				\\
				& on ImageNet, and then fine-tune on VOC & 
				\\
				\hline
			\end{tabular}
		}
	\end{center}
	\caption{ \small Results of semantic segmentation on the validation set of PASCAL VOC 2012. Note that ``VOC'' in ``fine-tune on VOC'' indicates the augmented training set of PASCAL VOC 2012.}
	\label{table: results of segmentation}
	\vspace{-0.1in}
\end{table}

\section{Conclusions}
\label{sec: conclusion}

In this work, we built a large-scale multi-label image database, dubbed {\it Tencent ML-Images}, including about 18M images and 11K categories. 
It is the largest-scale public multi-label image database until now. 
We presented the lage-scale visual presentation learning of deep convolutional neural networks on Tencent ML-Images, employing a distributed training framework with MPI and NCCL. 
A novel loss function was carefully designed to alleviate the side-effect of the severe class imbalance in the large-scale multi-label database. 
Extensive experiments of transfer learning to other visual tasks, including single-label image classification, object detection and semantic segmentation, verified that Tencent ML-Images is of very high quality and the pre-trained checkpoint has very good visual representation. 
We hope that this work could promote other visual tasks in the research and industry community. 
The Tencent ML-Images database, the complete code of data preparation, pre-training and fine-tuning, and the pre-trained and fine-tuned checkpoints of the ResNet-101 model have been released at {\it https://github.com/Tencent/tencent-ml-images}.

\bibliographystyle{IEEEtran}
\bibliography{bywu_bib}

\begin{thebibliography}{10}
\providecommand{\url}[1]{#1}
\csname url@samestyle\endcsname
\providecommand{\newblock}{\relax}
\providecommand{\bibinfo}[2]{#2}
\providecommand{\BIBentrySTDinterwordspacing}{\spaceskip=0pt\relax}
\providecommand{\BIBentryALTinterwordstretchfactor}{4}
\providecommand{\BIBentryALTinterwordspacing}{\spaceskip=\fontdimen2\font plus
\BIBentryALTinterwordstretchfactor\fontdimen3\font minus
  \fontdimen4\font\relax}
\providecommand{\BIBforeignlanguage}[2]{{%
\expandafter\ifx\csname l@#1\endcsname\relax
\typeout{** WARNING: IEEEtran.bst: No hyphenation pattern has been}%
\typeout{** loaded for the language `#1'. Using the pattern for}%
\typeout{** the default language instead.}%
\else
\language=\csname l@#1\endcsname
\fi
#2}}
\providecommand{\BIBdecl}{\relax}
\BIBdecl

\bibitem{alexnet-cnn-nips-2012}
A.~Krizhevsky, I.~Sutskever, and G.~E. Hinton, ``Imagenet classification with
  deep convolutional neural networks,'' in \emph{NIPS}, 2012, pp. 1097--1105.

\bibitem{imagenet-cvpr-2009}
J.~Deng, W.~Dong, R.~Socher, L.-J. Li, K.~Li, and L.~Fei-Fei, ``Imagenet: A
  large-scale hierarchical image database,'' in \emph{CVPR}, 2009, pp.
  248--255.

\bibitem{openimages}
I.~Krasin, T.~Duerig, N.~Alldrin, V.~Ferrari, S.~Abu-El-Haija, A.~Kuznetsova,
  H.~Rom, J.~Uijlings, S.~Popov, A.~Veit, S.~Belongie, V.~Gomes, A.~Gupta,
  C.~Sun, G.~Chechik, D.~Cai, Z.~Feng, D.~Narayanan, and K.~Murphy,
  ``Openimages: A public dataset for large-scale multi-label and multi-class
  image classification.'' \emph{Dataset available from
  https://github.com/openimages}, 2017.

\bibitem{kuznetsova2018open}
A.~Kuznetsova, H.~Rom, N.~Alldrin, J.~Uijlings, I.~Krasin, J.~Pont-Tuset,
  S.~Kamali, S.~Popov, M.~Malloci, T.~Duerig \emph{et~al.}, ``The open images
  dataset v4: Unified image classification, object detection, and visual
  relationship detection at scale,'' \emph{arXiv preprint arXiv:1811.00982},
  2018.

\bibitem{google-JFT300m-iccv-2017}
C.~Sun, A.~Shrivastava, S.~Singh, and A.~Gupta, ``Revisiting unreasonable
  effectiveness of data in deep learning era,'' in \emph{ICCV}, 2017, pp.
  843--852.

\bibitem{residual-he-cvpr-2016}
K.~He, X.~Zhang, S.~Ren, and J.~Sun, ``Deep residual learning for image
  recognition,'' in \emph{CVPR}, 2016, pp. 770--778.

\bibitem{wordnet-1998}
C.~Fellbaum, \emph{WordNet}.\hskip 1em plus 0.5em minus 0.4em\relax Wiley
  Online Library, 1998.

\bibitem{NCCL-2017}
NVIDIA, ``Nvidia collective communications library (nccl).''
  \emph{https://developer.nvidia.com/nccl}, 2017.

\bibitem{cifar10-2009}
A.~Krizhevsky and G.~Hinton, ``Learning multiple layers of features from tiny
  images,'' \emph{Technical report, University of Toronto}, 2009.

\bibitem{caltech-256}
G.~Griffin, A.~Holub, and P.~Perona, ``Caltech-256 object category dataset,''
  \emph{Technical Report 7694, Caltech}, 2007.

\bibitem{mnist-2010}
Y.~LeCun, L.~Bottou, Y.~Bengio, P.~Haffner \emph{et~al.}, ``Gradient-based
  learning applied to document recognition,'' \emph{Proceedings of the IEEE},
  vol.~86, no.~11, pp. 2278--2324, 1998.

\bibitem{webvision-2017}
W.~Li, L.~Wang, W.~Li, E.~Agustsson, and L.~Van~Gool, ``Webvision database:
  Visual learning and understanding from web data,'' \emph{arXiv preprint
  arXiv:1708.02862}, 2017.

\bibitem{sun-cvpr-2010}
J.~Xiao, J.~Hays, K.~A. Ehinger, A.~Oliva, and A.~Torralba, ``Sun database:
  Large-scale scene recognition from abbey to zoo,'' in \emph{CVPR}, 2010, pp.
  3485--3492.

\bibitem{places-2017}
B.~Zhou, A.~Lapedriza, A.~Khosla, A.~Oliva, and A.~Torralba, ``Places: A 10
  million image database for scene recognition,'' \emph{IEEE transactions on
  pattern analysis and machine intelligence}, vol.~40, no.~6, pp. 1452--1464,
  2017.

\bibitem{vgg-f-bmvc-2014}
K.~Chatfield, K.~Simonyan, A.~Vedaldi, and A.~Zisserman, ``Return of the devil
  in the details: Delving deep into convolutional nets,'' \emph{arXiv preprint
  arXiv:1405.3531}, 2014.

\bibitem{corel5k-eccv-2002}
P.~Duygulu, K.~Barnard, J.~F. de~Freitas, and D.~A. Forsyth, ``Object
  recognition as machine translation: Learning a lexicon for a fixed image
  vocabulary,'' in \emph{ECCV}, 2002, pp. 97--112.

\bibitem{espgame-2004}
L.~Von~Ahn and L.~Dabbish, ``Labeling images with a computer game,'' in
  \emph{Proceedings of the SIGCHI conference on Human factors in computing
  systems}, 2004, pp. 319--326.

\bibitem{iaprtc-12-data-2006}
M.~Grubinger, P.~Clough, H.~M{\"u}ller, and T.~Deselaers, ``The iapr tc-12
  benchmark: A new evaluation resource for visual information systems,'' in
  \emph{International Workshop OntoImage}, 2006, pp. 13--23.

\bibitem{nus-wide-civr09}
T.-S. Chua, J.~Tang, R.~Hong, H.~Li, Z.~Luo, and Y.-T. Zheng, ``Nus-wide: A
  real-world web image database from national university of singapore,'' in
  \emph{CIVR}, 2009, pp. 368--375.

\bibitem{mscoco-2014}
T.-Y. Lin, M.~Maire, S.~Belongie, J.~Hays, P.~Perona, D.~Ramanan,
  P.~Doll{\'a}r, and C.~L. Zitnick, ``Microsoft coco: Common objects in
  context,'' in \emph{ECCV}, 2014, pp. 740--755.

\bibitem{pascal-voc-2007}
M.~Everingham, L.~Van~Gool, C.~K. Williams, J.~Winn, and A.~Zisserman, ``The
  pascal visual object classes (voc) challenge,'' \emph{International journal
  of computer vision}, vol.~88, no.~2, pp. 303--338, 2010.

\bibitem{my-aaai-2016-imbalance}
B.~Wu, S.~Lyu, and B.~Ghanem, ``Constrained submodular minimization for missing
  labels and class imbalance in multi-label learning.'' in \emph{AAAI}, 2016,
  pp. 2229--2236.

\bibitem{my-icpr-2014}
B.~Wu, Z.~Liu, S.~Wang, B.-G. Hu, and Q.~Ji, ``Multi-label learning with
  missing labels,'' in \emph{ICPR}, 2014, pp. 1964--1968.

\bibitem{my-iccv-2015}
B.~Wu, S.~Lyu, and B.~Ghanem, ``Ml-mg: Multi-label learning with missing labels
  using a mixed graph,'' in \emph{ICCV}, 2015, pp. 4157--4165.

\bibitem{wu2015multi}
B.~Wu, S.~Lyu, B.-G. Hu, and Q.~Ji, ``Multi-label learning with missing labels
  for image annotation and facial action unit recognition,'' \emph{Pattern
  Recognition}, vol.~48, no.~7, pp. 2279--2289, 2015.

\bibitem{li2016facial}
Y.~Li, B.~Wu, B.~Ghanem, Y.~Zhao, H.~Yao, and Q.~Ji, ``Facial action unit
  recognition under incomplete data based on multi-label learning with missing
  labels,'' \emph{Pattern Recognition}, vol.~60, pp. 890--900, 2016.

\bibitem{my-ijcv-2018}
B.~Wu, F.~Jia, W.~Liu, B.~Ghanem, and S.~Lyu, ``Multi-label learning with
  missing labels using mixed dependency graphs,'' \emph{International Journal
  of Computer Vision}, vol. 126, no.~8, pp. 875--896, 2018.

\bibitem{openimages-noisy-learning-cvpr-2017}
A.~Veit, N.~Alldrin, G.~Chechik, I.~Krasin, A.~Gupta, and S.~Belongie,
  ``Learning from noisy large-scale datasets with minimal supervision,'' in
  \emph{CVPR}, 2017, pp. 839--847.

\bibitem{he2009learning}
H.~He and E.~A. Garcia, ``Learning from imbalanced data,'' \emph{IEEE
  Transactions on knowledge and data engineering}, vol.~21, no.~9, pp.
  1263--1284, 2009.

\bibitem{szegedy2015going}
C.~Szegedy, W.~Liu, Y.~Jia, P.~Sermanet, S.~Reed, D.~Anguelov, D.~Erhan,
  V.~Vanhoucke, and A.~Rabinovich, ``Going deeper with convolutions,'' in
  \emph{CVPR}, 2015, pp. 1--9.

\bibitem{back-propagation-hinton-1986}
D.~E. Rumelhart, G.~E. Hinton, and R.~J. Williams, ``Learning representations
  by back-propagating errors,'' \emph{Cognitive modeling}, vol.~5, no.~3, p.~1,
  1988.

\bibitem{warmup-2017}
P.~Goyal, P.~Doll{\'a}r, R.~Girshick, P.~Noordhuis, L.~Wesolowski, A.~Kyrola,
  A.~Tulloch, Y.~Jia, and K.~He, ``Accurate, large minibatch sgd: training
  imagenet in 1 hour,'' \emph{arXiv preprint arXiv:1706.02677}, 2017.

\bibitem{all-reduce-2009}
P.~Patarasuk and X.~Yuan, ``Bandwidth optimal all-reduce algorithms for
  clusters of workstations,'' \emph{Journal of Parallel and Distributed
  Computing}, vol.~69, no.~2, pp. 117--124, 2009.

\bibitem{zhang2013review}
M.-L. Zhang and Z.-H. Zhou, ``A review on multi-label learning algorithms,''
  \emph{IEEE transactions on knowledge and data engineering}, vol.~26, no.~8,
  pp. 1819--1837, 2013.

\bibitem{resnet-v2-eccv-2016}
K.~He, X.~Zhang, S.~Ren, and J.~Sun, ``Identity mappings in deep residual
  networks,'' in \emph{ECCV}, 2016, pp. 630--645.

\bibitem{yolo9000}
J.~Redmon and A.~Farhadi, ``Yolo9000: Better, faster, stronger,'' in
  \emph{CVPR}, 2017, pp. 6517--6525.

\bibitem{everingham2015pascal}
M.~Everingham, S.~A. Eslami, L.~Van~Gool, C.~K. Williams, J.~Winn, and
  A.~Zisserman, ``The pascal visual object classes challenge: A
  retrospective,'' \emph{International journal of computer vision}, vol. 111,
  no.~1, pp. 98--136, 2015.

\bibitem{cnn-object-detection-nips-2015}
S.~Ren, K.~He, R.~Girshick, and J.~Sun, ``Faster r-cnn: Towards real-time
  object detection with region proposal networks,'' in \emph{NIPS}, 2015, pp.
  91--99.

\bibitem{chen2017spatial}
X.~Chen and A.~Gupta, ``Spatial memory for context reasoning in object
  detection,'' in \emph{ICCV}, 2017, pp. 4086--4096.

\bibitem{chen2018deeplab}
L.-C. Chen, G.~Papandreou, I.~Kokkinos, K.~Murphy, and A.~L. Yuille, ``Deeplab:
  Semantic image segmentation with deep convolutional nets, atrous convolution,
  and fully connected crfs,'' \emph{IEEE transactions on pattern analysis and
  machine intelligence}, vol.~40, no.~4, pp. 834--848, 2018.

\end{thebibliography}

\end{document}